\documentclass{article}\usepackage[]{graphicx}\usepackage[]{color}
\makeatletter
\def\maxwidth{ %
  \ifdim\Gin@nat@width>\linewidth
    \linewidth
  \else
    \Gin@nat@width
  \fi
}
\makeatother

\definecolor{fgcolor}{rgb}{0.345, 0.345, 0.345}

\usepackage{framed}
\makeatletter
 {\par\unskip\endMakeFramed%
 \at@end@of@kframe}
\makeatother

\definecolor{shadecolor}{rgb}{.97, .97, .97}
\definecolor{messagecolor}{rgb}{0, 0, 0}
\definecolor{warningcolor}{rgb}{1, 0, 1}
\definecolor{errorcolor}{rgb}{1, 0, 0}
\newenvironment{knitrout}{}{} 

\usepackage{alltt}

\usepackage{booktabs}
\usepackage{hyperref}
\usepackage{rotating}
\usepackage{float}
\usepackage{caption}

\usepackage{listings}
\title{ICON Challenge on Algorithm Selection\\[1em]
\Large\url{http://challenge.icon-fet.eu}}

\author{Lars Kotthoff}
\IfFileExists{upquote.sty}{\usepackage{upquote}}{}
\begin{document}

\maketitle

\section{Submissions}

The challenge received a total of 8 submissions from 4 different groups of
researchers comprising 15 people. Participants were based in 4 different
countries on 2 continents. Table~\ref{tab:subs} gives an overview of all
submissions.

\begin{table}
\begin{center}
\begin{tabular}{lll}
\toprule
System name & Presolving/feature selection used?\\
\midrule
ASAP\_kNN & no\\
ASAP\_RF & no\\
autofolio & yes\\
flexfolio-schedules & yes\\
sunny & no\\
sunny-presolv & yes\\
zilla & yes\\
zillafolio & yes\\
\bottomrule
\end{tabular}
\caption{Systems submitted to the challenge.}\label{tab:subs}
\end{center}
\end{table}

All submissions were submitted for evaluation on all ASlib scenarios. Most
systems used a presolver and specified a subset of features to use for each
scenario.

\section{Evaluation}

The evaluation was performed as follows. For each scenario, 10 bootstrap
samplings of the entire data were used to create 10 different train/test splits.
No stratification was used. The training part was left unmodified. For the test
part, algorithm performances were set to 0 and runstatus to ``ok'' for all
algorithms and all instances -- the ASlib specification requires algorithm
performance data to be part of a scenario. A \verb+cv.arff+ file was generated
for both training and testing with 10 folds and the instances assigned to folds
by the order in which they appeared in the original scenario.

For systems that specified a presolver, the instances that were solved by the
presolver within the specified time were removed from the training set. If a
subset of features was specified, only these features (and only the costs
associated with these features) were left in both training and test set, with
all other feature values removed.

Each system was trained on each train scenario and predicted on each test
scenario. In total, 130 evaluations (10 for each of the 13 scenarios) per
submitted system were performed. The total CPU time spent was 
4685.11 hours.

\medskip

The predictions were evaluated as follows. If a presolver was specified, it was
``run'' for the specified time. If the instance was solved within this time, the
time to solve the instance was taken as the performance on that instance and the
instance recorded as solved.

Otherwise, the time limit given for the presolving run was added to the time
required to compute all features specified for the particular scenario. For any
instances that were solved during feature computation, the instance was recorded
as solved at this point and the time for the presolving run plus feature
computation recorded as the performance. The misclassification penalty was set
to 0 in this case regardless of the performance of the best solver.

For instances not solved during feature computation, the solvers specified in
the prediction schedule of the system were ``run''. For each instance, the
predicted solvers were ordered by the \verb+runID+ specified. If a run was
unable to solve an instance, the smaller of time the schedule specified to run
it for and the time it actually took to run on the instance was added to the
total. If a run solved the respective instance, the actual time required by the
algorithm was added to the total and the instance recorded as solved. If the
total time exceeded the time limit for the scenario, an instance was recorded as
not solved.

Each system was evaluated in terms of mean PAR10 score, mean misclassification
penalty, and mean number of instances solved for each of the 130 evaluations on
each scenario and split.

To facilitate comparison of the different measures across the different
scenarios, all measures were normalised by the performance of the virtual best
(VBS) and the single best (SB) solver. The single best solver was determined as
the solver with the smallest overall runtime across all instances.
Equation~\ref{eq:norm} defines the normalisation of a score $s$.

\begin{equation}\label{eq:norm}
s_{norm} = \frac{s - s_{VBS}}{s_{SB} - s_{VBS}}
\end{equation}

This normalises the score to the interval 0 (VBS) to 1 (SB), with smaller values
being better. The number denotes how much of the gap between single best and
virtual best solver was left by the system.

\medskip

To determine the overall winner, the mean across all of the normalised
measurements was taken. For each submitted system, 390 scores were taken into
account for this (13 scenarios times 10 splits times 3 measures).

\section{Results}

Table~\ref{tab:ranks} shows the final ranking. The first and second placed
entries are very close. All systems perform well on average, closing more than
half of the gap between virtual and single best solver.

For comparison, we show three other systems. Autofolio-48 is identical to
Autofolio, but was allowed 48 hours training time to assess the impact of
additional exploration of the hyperparameter space. Llama-regrPairs and
llama-regr are simple llama models (see Appendix~\ref{app:llama}).

\begin{table}[ht]
\centering
\begin{tabular}{rlr}
  \toprule
 & System & Average total score \\ 
  \midrule
1 & zilla & 0.36603 \\ 
  2 & zillafolio & 0.37021 \\ 
  3 & autofolio-48 & 0.37500 \\ 
  4 & autofolio & 0.39083 \\ 
  5 & llama-regrPairs & 0.39501 \\ 
  6 & ASAP\_RF & 0.41603 \\ 
  7 & ASAP\_kNN & 0.42318 \\ 
  8 & llama-regr & 0.42515 \\ 
  9 & flexfolio-schedules & 0.44251 \\ 
  10 & sunny & 0.48259 \\ 
  11 & sunny-presolv & 0.48488 \\ 
   \bottomrule
\end{tabular}
\caption{Final ranking.} 
\label{tab:ranks}
\end{table}

To assess how significant the difference are and how stable the ranking is, we
took 1\,000 bootstrap samples from the scenario-split combinations and computed
the scores and ranks on each of them. The mean average of the total score
averages over the bootstrap samples and the confidence intervals are show in
Table~\ref{tab:ranksboot}.

\begin{table}[ht]
\centering
\begin{tabular}{rlrrr}
  \toprule
 & System & Average total score & 95\% CI upper & 95\% CI lower \\ 
  \midrule
1 & zilla & 0.36631 & 0.36735 & 0.36527 \\ 
  2 & zillafolio & 0.37039 & 0.37151 & 0.36928 \\ 
  3 & autofolio-48 & 0.37557 & 0.37671 & 0.37442 \\ 
  4 & autofolio & 0.39106 & 0.39224 & 0.38988 \\ 
  5 & llama-regrPairs & 0.39550 & 0.39669 & 0.39432 \\ 
  6 & ASAP\_RF & 0.41656 & 0.41801 & 0.41511 \\ 
  7 & ASAP\_kNN & 0.42383 & 0.42528 & 0.42237 \\ 
  8 & llama-regr & 0.42541 & 0.42668 & 0.42414 \\ 
  9 & flexfolio-schedules & 0.44278 & 0.44426 & 0.44129 \\ 
  10 & sunny & 0.48298 & 0.48454 & 0.48141 \\ 
  11 & sunny-presolv & 0.48514 & 0.48667 & 0.48361 \\ 
   \bottomrule
\end{tabular}
\caption{Final ranking, bootstrapped.} 
\label{tab:ranksboot}
\end{table}

The ranking is the same as the final ranking in Table~\ref{tab:ranks}. The
confidence intervals show that the rankings are relatively stable.

\subsection{Winner -- zilla}

The winner of the ICON Challenge on Algorithm Selection is zilla by Chris
Cameron, Alex Fr\'echette, Holger Hoos, Frank Hutter, and Kevin Leyton-Brown.

\subsection{Honourable mention -- ASAP\_RF}

ASAP\_RF by Fran\c{}cois Gonard, Marc Schoenauer, and Mich\`ele Sebag receives an
honourable mention as a submission that has not been described in the literature
before and showed respectable performance, beating all other approaches in some
cases.

\subsection{Alternative rank aggregations}

An alternative (and probably fairer) way of determining the winner is to see the
ranking of systems induced by each measure on each split of each scenario as a
ballot (for a total of 260 ballots) and aggregate the ranks in those ballots.
Here, we optimise the aggregated Spearman coefficient between candidate rankings
and ballot rankings. That is, the final ranking has the optimal Spearman
coefficient with respect to the ballots.

Table~\ref{tab:aggranks} shows the aggregated ranks. Now autofolio is in second
position.

\begin{table}[ht]
\centering
\begin{tabular}{rl}
  \toprule
 & System \\ 
  \midrule
1 & zilla \\ 
  2 & autofolio \\ 
  3 & autofolio-48 \\ 
  4 & zillafolio \\ 
  5 & llama-regrPairs \\ 
  6 & ASAP\_RF \\ 
  7 & ASAP\_kNN \\ 
  8 & llama-regr \\ 
  9 & flexfolio-schedules \\ 
  10 & sunny \\ 
  11 & sunny-presolv \\ 
   \bottomrule
\end{tabular}
\caption{Aggregated ranks.} 
\label{tab:aggranks}
\end{table}

There are significant changes however when averaging the performance across all
measures, splits, and scenarios by median rather than mean.
Table~\ref{tab:medranks} shows this ranking. Zilla is now in second position,
beat by ASAP\_RF.

\begin{table}[ht]
\centering
\begin{tabular}{rlr}
  \toprule
 & System & Median total score \\ 
  \midrule
1 & ASAP\_RF & 0.28566 \\ 
  2 & zilla & 0.29262 \\ 
  3 & autofolio-48 & 0.29669 \\ 
  4 & autofolio & 0.30043 \\ 
  5 & llama-regrPairs & 0.30358 \\ 
  6 & zillafolio & 0.30714 \\ 
  7 & ASAP\_kNN & 0.30858 \\ 
  8 & llama-regr & 0.32373 \\ 
  9 & flexfolio-schedules & 0.33305 \\ 
  10 & sunny & 0.37355 \\ 
  11 & sunny-presolv & 0.41260 \\ 
   \bottomrule
\end{tabular}
\caption{Ranking by median.} 
\label{tab:medranks}
\end{table}

\subsection{Detailed results}

Tables~\ref{tab:rankspar10} through~\ref{tab:rankssolved} show the rankings by
mean score across all splits and scenarios, but separately for each measure.

\begin{table}[ht]
\centering
\begin{tabular}{rlr}
  \toprule
 & System & Mean PAR10 score \\ 
  \midrule
1 & autofolio-48 & 0.33383 \\ 
  2 & autofolio & 0.34104 \\ 
  3 & zilla & 0.34414 \\ 
  4 & zillafolio & 0.34553 \\ 
  5 & llama-regrPairs & 0.37496 \\ 
  6 & ASAP\_RF & 0.37749 \\ 
  7 & ASAP\_kNN & 0.38658 \\ 
  8 & flexfolio-schedules & 0.39518 \\ 
  9 & llama-regr & 0.40749 \\ 
  10 & sunny & 0.46144 \\ 
  11 & sunny-presolv & 0.46657 \\ 
   \bottomrule
\end{tabular}
\caption{Ranking by PAR10.} 
\label{tab:rankspar10}
\end{table}
\begin{table}[ht]
\centering
\begin{tabular}{rlr}
  \toprule
 & System & Mean misclassification penalty \\ 
  \midrule
1 & zilla & 0.41874 \\ 
  2 & zillafolio & 0.43001 \\ 
  3 & llama-regrPairs & 0.44447 \\ 
  4 & llama-regr & 0.46856 \\ 
  5 & autofolio-48 & 0.47449 \\ 
  6 & ASAP\_RF & 0.51015 \\ 
  7 & autofolio & 0.51131 \\ 
  8 & ASAP\_kNN & 0.51263 \\ 
  9 & sunny-presolv & 0.52817 \\ 
  10 & sunny & 0.53324 \\ 
  11 & flexfolio-schedules & 0.55644 \\ 
   \bottomrule
\end{tabular}
\caption{Ranking by misclassification penalty.} 
\label{tab:ranksmcp}
\end{table}
\begin{table}[ht]
\centering
\begin{tabular}{rlr}
  \toprule
 & System & Mean number of instances solved \\ 
  \midrule
1 & autofolio-48 & 0.31668 \\ 
  2 & autofolio & 0.32015 \\ 
  3 & zillafolio & 0.33509 \\ 
  4 & zilla & 0.33522 \\ 
  5 & ASAP\_RF & 0.36045 \\ 
  6 & llama-regrPairs & 0.36559 \\ 
  7 & ASAP\_kNN & 0.37035 \\ 
  8 & flexfolio-schedules & 0.37592 \\ 
  9 & llama-regr & 0.39941 \\ 
  10 & sunny & 0.45309 \\ 
  11 & sunny-presolv & 0.45990 \\ 
   \bottomrule
\end{tabular}
\caption{Ranking by number of instances solved.} 
\label{tab:rankssolved}
\end{table}

\clearpage

Table~\ref{tab:ranksscen} shows the ranks for the different scenarios for all
systems by mean across all measures and splits.

\begin{table}[ht]
\centering
\begin{tabular}{llllllllllll}
  \toprule
scenario & \begin{sideways}ASAP\_kNN\end{sideways} & \begin{sideways}ASAP\_RF\end{sideways} & \begin{sideways}autofolio-48\end{sideways} & \begin{sideways}autofolio\end{sideways} & \begin{sideways}flexfolio-schedules\end{sideways} & \begin{sideways}llama-regr\end{sideways} & \begin{sideways}llama-regrPairs\end{sideways} & \begin{sideways}sunny\end{sideways} & \begin{sideways}sunny-presolv\end{sideways} & \begin{sideways}zilla\end{sideways} & \begin{sideways}zillafolio\end{sideways} \\ 
  \midrule
ASP-POTASSCO & 10 & 5 & 2 & 6 & 7 & 3 & \textcolor{blue}{\textbf{1}} & \textcolor{red}{\textbf{11}} & 9 & 8 & 4 \\ 
  CSP-2010 & 7 & \textcolor{blue}{\textbf{1}} & 10 & 9 & 6 & 5 & 3 & 8 & \textcolor{red}{\textbf{11}} & 4 & 2 \\ 
  MAXSAT12-PMS & 4 & 6 & 10 & \textcolor{red}{\textbf{11}} & 2 & 3 & \textcolor{blue}{\textbf{1}} & 5 & 8 & 7 & 9 \\ 
  PREMARSHALLING-ASTAR-2013 & 6 & 4 & 7 & 5 & 2 & 10 & \textcolor{red}{\textbf{11}} & 3 & \textcolor{blue}{\textbf{1}} & 8 & 9 \\ 
  PROTEUS-2014 & 5 & 4 & 7 & 6 & \textcolor{blue}{\textbf{1}} & \textcolor{red}{\textbf{11}} & 10 & 3 & 2 & 9 & 8 \\ 
  QBF-2011 & \textcolor{blue}{\textbf{1}} & 2 & 9 & 8 & 3 & 6 & 4 & 5 & 7 & 10 & \textcolor{red}{\textbf{11}} \\ 
  SAT11-HAND & 3 & 6 & \textcolor{blue}{\textbf{1}} & 7 & 8 & 9 & 5 & \textcolor{red}{\textbf{11}} & 10 & 4 & 2 \\ 
  SAT11-INDU & 6 & 8 & 2 & 3 & 10 & 5 & \textcolor{blue}{\textbf{1}} & 9 & \textcolor{red}{\textbf{11}} & 4 & 7 \\ 
  SAT11-RAND & 6 & 7 & \textcolor{blue}{\textbf{1}} & 2 & \textcolor{red}{\textbf{11}} & 10 & 8 & 9 & 5 & 3 & 4 \\ 
  SAT12-ALL & 7 & 8 & 2 & 3 & 9 & 6 & 5 & 10 & \textcolor{red}{\textbf{11}} & \textcolor{blue}{\textbf{1}} & 4 \\ 
  SAT12-HAND & 7 & 8 & 4 & \textcolor{blue}{\textbf{1}} & 9 & 6 & 3 & 10 & \textcolor{red}{\textbf{11}} & 2 & 5 \\ 
  SAT12-INDU & 8 & 9 & \textcolor{blue}{\textbf{1}} & 3 & 7 & 6 & 5 & \textcolor{red}{\textbf{11}} & 10 & 4 & 2 \\ 
  SAT12-RAND & 10 & 8 & 3 & 4 & 9 & 6 & 5 & \textcolor{red}{\textbf{11}} & 7 & \textcolor{blue}{\textbf{1}} & 2 \\ 
   \bottomrule
\end{tabular}
\caption{Ranks by scenario.} 
\label{tab:ranksscen}
\end{table}

Figures~\ref{fig:par10} through~\ref{fig:solved} give a more detailed overview
of the performance of the systems on the different scenarios. The colour of each
boxplot denotes the system, the mean performance of which is shown in the
legend (this corresponds to the number in the respective table above). The
boxplot shows the variation of performance across the 10 different splits for
each scenario. The solid black line denotes the performance of the single best
solver; anything above is worse.

\begin{knitrout}
\definecolor{shadecolor}{rgb}{0.969, 0.969, 0.969}\color{fgcolor}\begin{figure}
\includegraphics[width=\maxwidth]{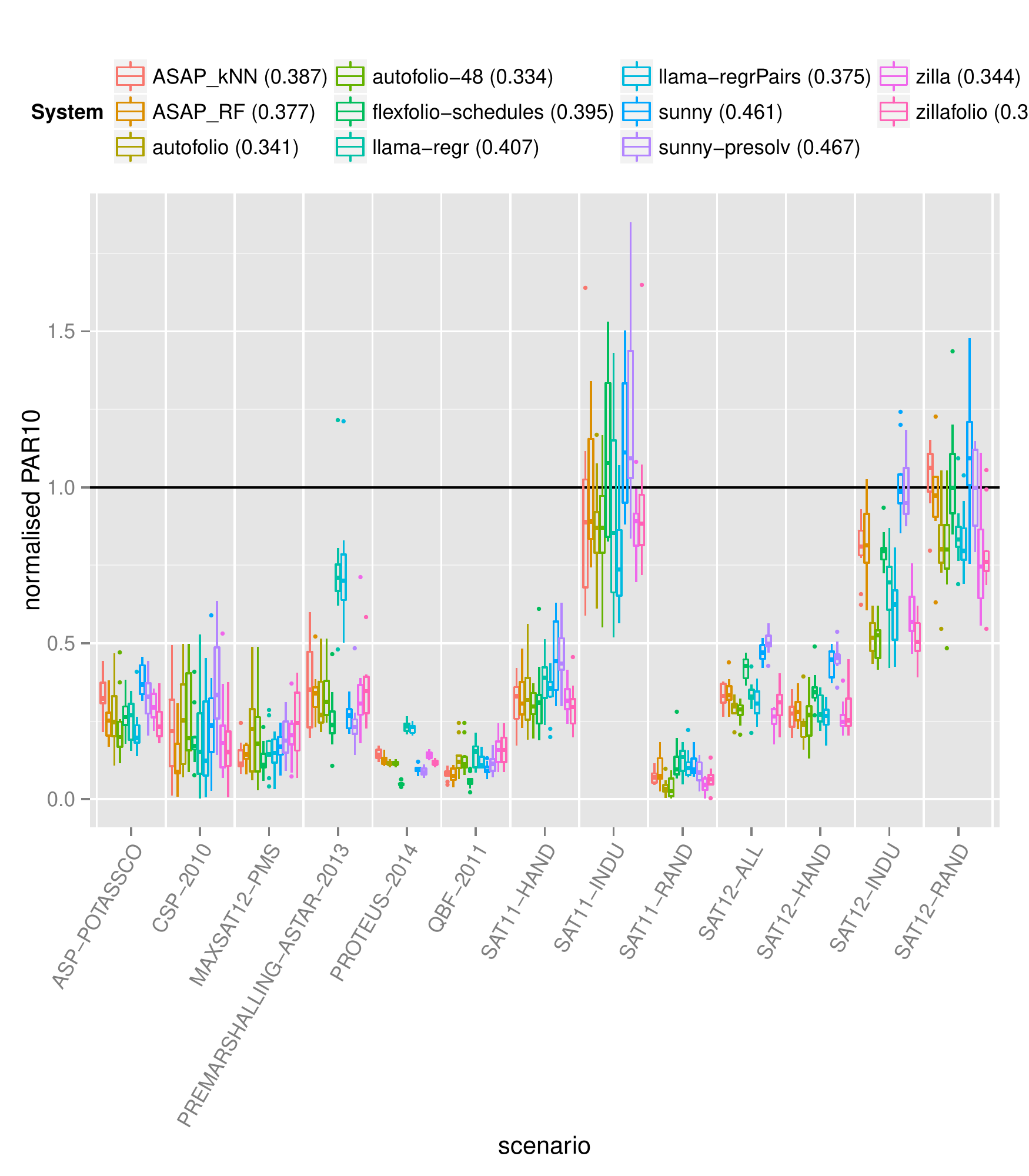} \caption[PAR10 scores]{PAR10 scores.}\label{fig:par10}
\end{figure}

\end{knitrout}

\begin{knitrout}
\definecolor{shadecolor}{rgb}{0.969, 0.969, 0.969}\color{fgcolor}\begin{figure}
\includegraphics[width=\maxwidth]{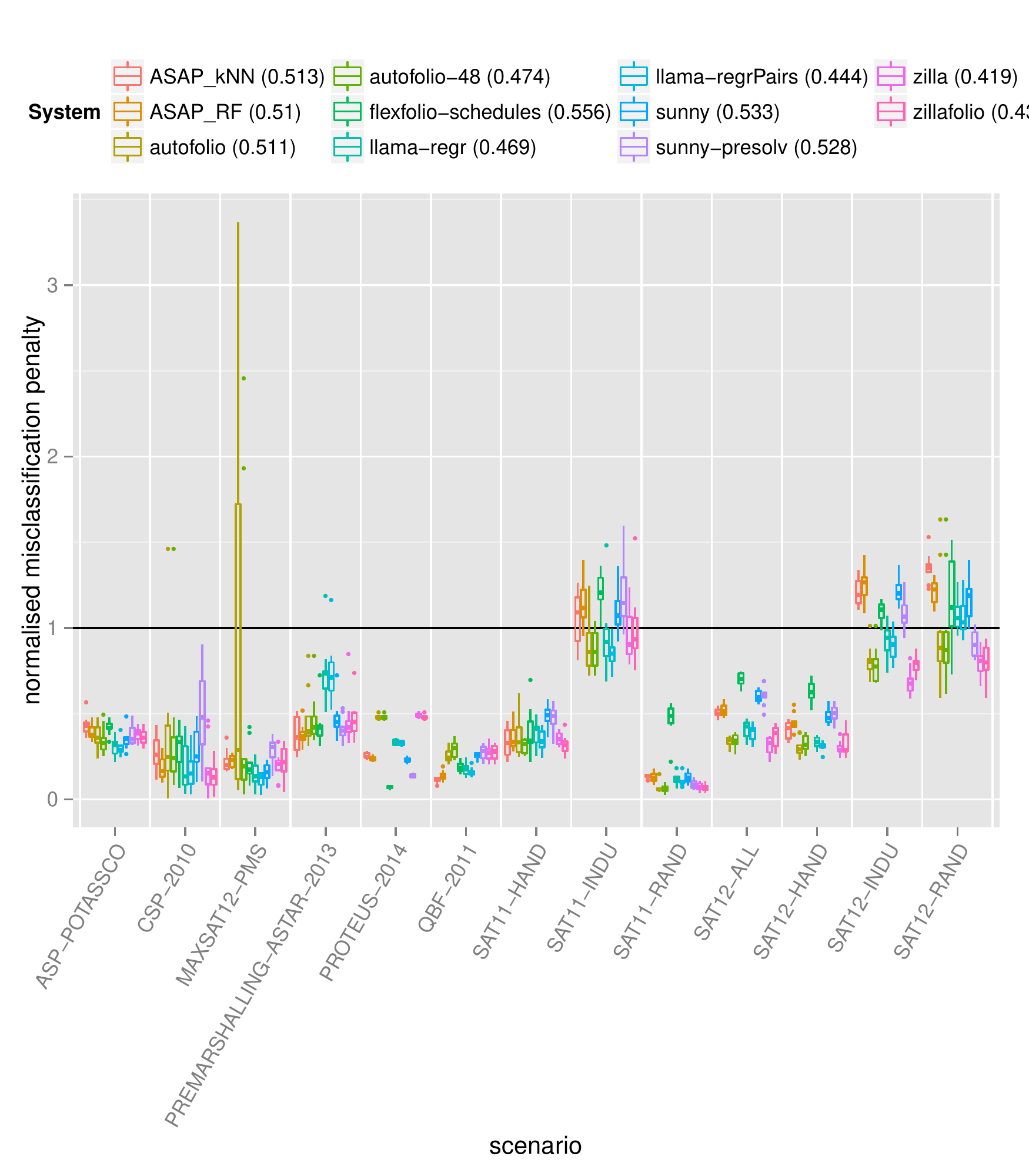} \caption[Misclassification penalty]{Misclassification penalty.}\label{fig:mcp}
\end{figure}

\end{knitrout}

\begin{knitrout}
\definecolor{shadecolor}{rgb}{0.969, 0.969, 0.969}\color{fgcolor}\begin{figure}
\includegraphics[width=\maxwidth]{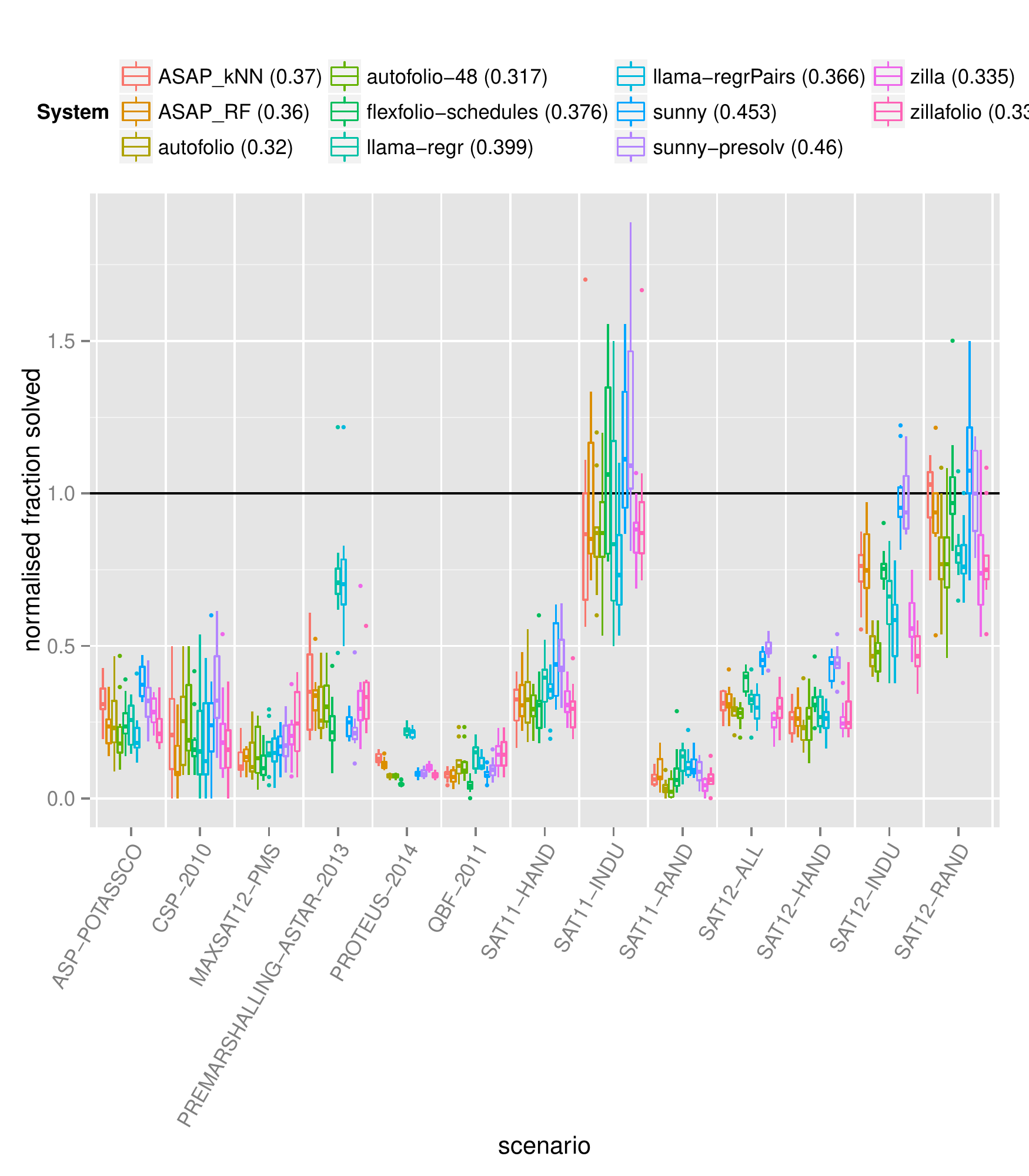} \caption[Instances solved]{Instances solved.}\label{fig:solved}
\end{figure}

\end{knitrout}

Two of the SAT scenarios are hard for all systems in the sense that the
performance they deliver on at least one of the splits is worse than the
performance of the single best solver. For most other scenarios, using any
algorithm selection system gives a significant performance improvement compared
to the single best solver though.

\subsection{Time required to run}

The time required to train the models and make the predictions varied
significantly across systems and scenarios, with some completing in minutes and
others requiring hours. Figure~\ref{fig:time} presents a summary.

\begin{knitrout}
\definecolor{shadecolor}{rgb}{0.969, 0.969, 0.969}\color{fgcolor}\begin{figure}
\includegraphics[width=\maxwidth]{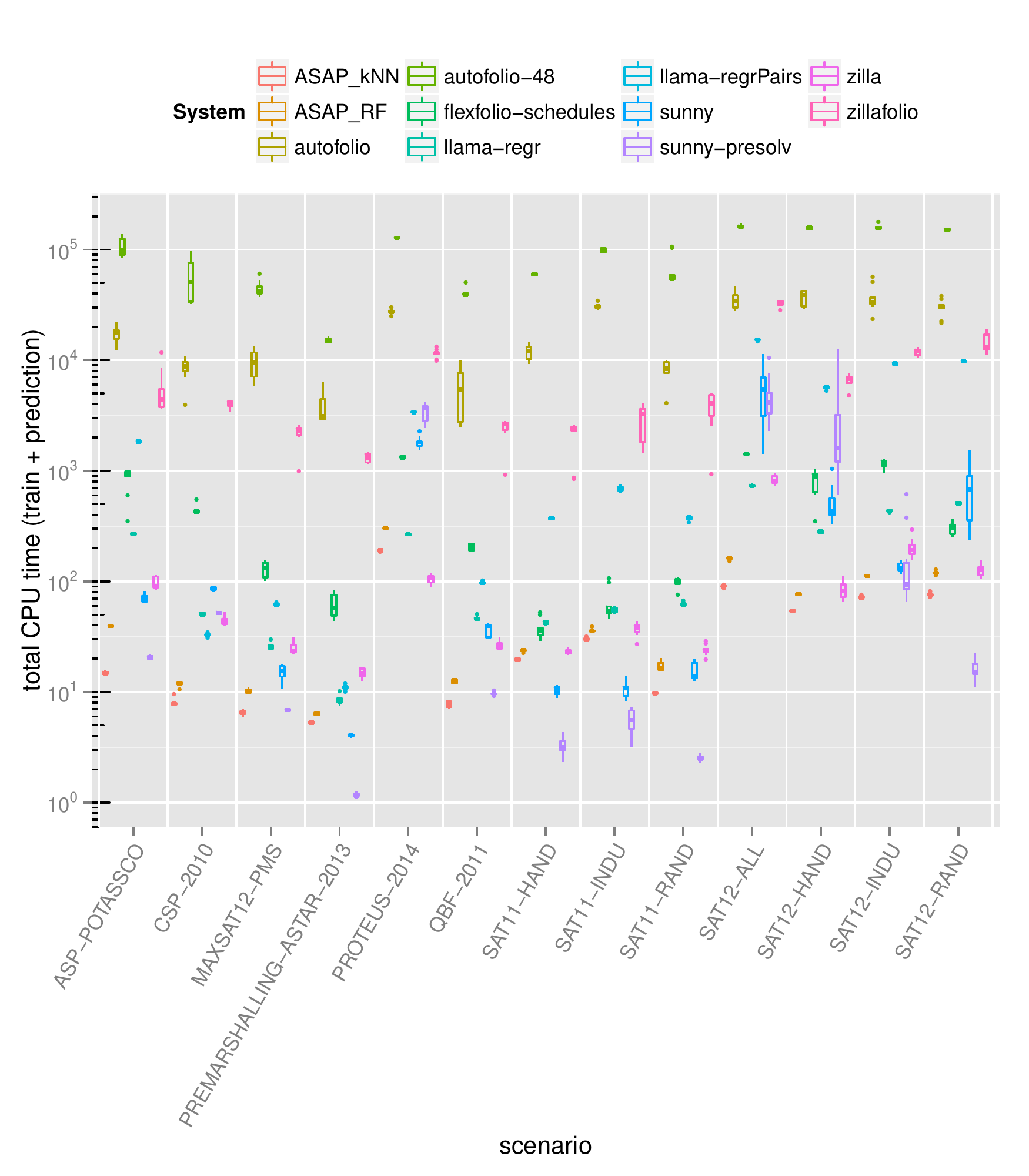} \caption[Train + prediction time]{Train + prediction time.}\label{fig:time}
\end{figure}

\end{knitrout}

\section{Acknowledgements}

We would like to thank all the participants for taking the time to prepare
submissions and their help in getting them to run; in alphabetical order:
Alex Fr\'echette,
Chris Cameron,
David Bergdoll,
Fabio Biselli,
Fran\c{}cois Gonard,
Frank Hutter,
Holger Hoos,
Jacopo Mauro,
Kevin Leyton-Brown,
Marc Schoenauer,
Marius Lindauer,
Mich\`ele Sebag,
Roberto Amadini,
Tong Liu, and
Torsten Schaub.
We thank Barry Hurley for setting up and maintaining the submission website and
Luc De Raedt, Siegfried Nijssen, Benjamin Negrevergne, Behrouz Babaki, Bernd
Bischl, and Marius Lindauer for feedback on the design of the challenge.

All data, code and results from the challenge are available at
\url{http://4c.ucc.ie/~larsko/downloads/challenge.tar.gz}.

\appendix

\section{Llama models used for comparison}\label{app:llama}

\subsection{llama-regrPairs}

\begin{lstlisting}
suppressMessages({
library(optparse)
library(aslib)
library(llama)
library(plyr)
})

ol = list(make_option(c("-t", "--train"), help = "AS scenario for training"),
    make_option(c("-p", "--prediction"), help = "AS scenario for predictions"))
opts = parse_args(OptionParser(option_list = ol))

suppressWarnings({trainAS = parseASScenario(opts$train)})
suppressWarnings({suppressMessages({ldf = convertToLlama(trainAS)})})
suppressWarnings({testAS = parseASScenario(opts$prediction)})
suppressWarnings({suppressMessages({ldft = convertToLlama(testAS)})})

# some features are removed by the conversion, make sure that we use only the
# intersection
feats = intersect(ldf$features, ldft$features)
ldf$features = feats
ldft$features = feats

tt = trainTest(ldf)
model = regressionPairs(makeLearner("regr.randomForest"), tt)

preds = model$predictor(ldft$data[,feats])

sched = ddply(preds, c("id"), function(ss) {
    data.frame(instanceID = testAS$feature.values[ss$id[1],"instance_id"],
        runID = 1,
        solver = ss$algorithm[1],
        timeLimit = testAS$desc$algorithm_cutoff_time)
})

write.csv(sched[,c("instanceID", "runID", "solver", "timeLimit")], file = stdout(), quote = FALSE, row.names = FALSE)
\end{lstlisting}

\subsection{llama-regr}

\begin{lstlisting}
suppressMessages({
library(optparse)
library(aslib)
library(llama)
library(plyr)
})

ol = list(make_option(c("-t", "--train"), help = "AS scenario for training"),
    make_option(c("-p", "--prediction"), help = "AS scenario for predictions"))
opts = parse_args(OptionParser(option_list = ol))

suppressWarnings({trainAS = parseASScenario(opts$train)})
suppressWarnings({suppressMessages({ldf = convertToLlama(trainAS)})})
suppressWarnings({testAS = parseASScenario(opts$prediction)})
suppressWarnings({suppressMessages({ldft = convertToLlama(testAS)})})

# some features are removed by the conversion, make sure that we use only the
# intersection
feats = intersect(ldf$features, ldft$features)
ldf$features = feats
ldft$features = feats

tt = trainTest(ldf)
model = regression(makeLearner("regr.randomForest"), tt)

preds = model$predictor(ldft$data[,feats])

sched = ddply(preds, c("id"), function(ss) {
    data.frame(instanceID = testAS$feature.values[ss$id[1],"instance_id"],
        runID = 1,
        solver = ss$algorithm[1],
        timeLimit = testAS$desc$algorithm_cutoff_time)
})

write.csv(sched[,c("instanceID", "runID", "solver", "timeLimit")], file = stdout(), quote = FALSE, row.names = FALSE)
\end{lstlisting}

\nocite{*}
\bibliographystyle{plain}
\bibliography{\jobname}

\end{document}